\documentclass[11pt,a4paper]{article}
\usepackage[hyperref]{acl2020}
\usepackage{times}
\usepackage{latexsym}

\usepackage{microtype}

\aclfinalcopy 
\def\aclpaperid{56} 

\usepackage{amsmath,amsfonts,bm}
\usepackage{booktabs} 
\usepackage{url}
\usepackage{graphicx}
\usepackage{array}
\usepackage{color}
\usepackage{makecell}
\usepackage{multirow}
\usepackage{xparse}

\newcommand{\ie}{\textit{i.e.}} 
\newcommand{\eg}{\textit{e.g.}} 

\newcommand{\start}[1]{\vspace{0.5mm}\noindent{{\bf #1}}}
\newcolumntype{P}[1]{>{\centering\arraybackslash}p{#1}}

\definecolor{gred}{RGB}{219,68,55}
\definecolor{gblue}{RGB}{66,133,244}
\definecolor{gyellow}{RGB}{244,180,0}
\definecolor{ggreen}{RGB}{15,157,88}
\definecolor{ggrey}{RGB}{115,115,115}

\newcommand{\colorG}[1]{\textcolor{ggreen}{\textbf{#1}}}
\newcommand{\colorB}[1]{\textcolor{gblue}{\textbf{#1}}}

\title{Facet-Aware Evaluation for Extractive Summarization}

\author{Yuning Mao$^{1}$, Liyuan Liu$^{1}$, Qi Zhu$^{1}$, Xiang Ren$^2$, Jiawei Han$^1$ \\
$^1$Department of Computer Science, University of Illinois at Urbana-Champaign, IL, USA \\
$^2$Department of Computer Science, University of Southern California, CA, USA\\
$^1$\{yuningm2, ll2, qiz3, hanj\}@illinois.edu $\quad$ $^2$xiangren@usc.edu
}

\date{}

\begin{document}
\maketitle

\begin{abstract}
    Commonly adopted metrics for extractive summarization focus on lexical overlap at the token level.
    In this paper, we present a facet-aware evaluation setup for better assessment of the information coverage in extracted summaries.
    Specifically, we treat each sentence in the reference summary as a \textit{facet}, identify the sentences in the document that express the semantics of each facet as \textit{support sentences} of the facet, and automatically evaluate extractive summarization methods by comparing the indices of extracted sentences and support sentences of all the facets in the reference summary.
    To facilitate this new evaluation setup, we construct an extractive version of the CNN/Daily Mail dataset and perform a thorough quantitative investigation, through which
    we demonstrate that facet-aware evaluation manifests better correlation with human judgment than ROUGE, enables fine-grained evaluation as well as comparative analysis, and reveals valuable insights of state-of-the-art summarization methods.\footnote{Data can be found at \url{https://github.com/morningmoni/FAR}.}
\end{abstract}

\section{Introduction}
Text summarization has enjoyed increasing popularity due to its wide applications, whereas the evaluation of text summarization remains challenging and controversial.
The most commonly used evaluation metric of summarization is lexical overlap,
\ie, ROUGE~\cite{lin-2004-rouge},
which regards the system and reference summaries as sequences of tokens and measures their n-gram overlap.

However, recent studies~\cite{paulus2017deep,schluter-2017-limits,kryscinski2019neural} reveal the limitations
of ROUGE and find that in many cases, it fails to reach consensus with human judgment.
Since lexical overlap only captures information coverage at the surface (token) level, ROUGE favors system summaries that share more tokens with the reference summaries. Nevertheless, such summaries may not always convey the desired semantics.
For example, in Table~\ref{table_bad}, 
the document sentence with the highest ROUGE score has more lexical overlap  but expresses rather different semantic meaning. 
In contrast, the sentence manually extracted from the document by our annotators, which conveys similar semantics, is over-penalized as it involves other details or uses alternative words.

\begin{table}[t]
    \small
        \vspace{-0.6cm}
        \scalebox{.8}{
        \begin{tabular}{p{9.2cm}}
            \toprule
             \textbf{Reference}: Three people in \colorG{Kansas} have died from a \colorG{listeria outbreak}.  \\
             \textbf{Lexical Overlap}: But they did not appear identical to \colorB{listeria} samples taken from patients infected in the \colorB{Kansas outbreak}. (\textit{\textbf{ROUGE-1 F1=37.0, multiple token matches but totally different semantics}}) \\
             \textbf{Manual Extract}: Five people were infected and \colorB{three died} in the past year \colorB{in Kansas from listeria} that might be linked to blue bell creameries products, according to the CDC. (\textit{\textbf{ROUGE-1 F1=36.9, semantics covered but lower ROUGE due to the presence of other details)}}\\

             \midrule
             \textbf{Reference}: Chelsea boss \colorG{Jose Mourinho} and United manager \colorG{Louis van Gaal} are pals. \\
             \textbf{Lexical Overlap}: Gary Neville believes \colorB{Louis van Gaal}'s greatest achievement as a football manager is the making of \colorB{Jose Mourinho}. \\
             \textbf{Manual Extract}: The duo have been friends since they first worked together at Barcelona in 1997 where they enjoyed a successful relationship at the Camp Nou. (\textit{\textbf{ROUGE Recall/F1=0, no lexical overlap at all}})\\
            \bottomrule
        \end{tabular}
        }
    \vspace*{-.1cm}
    \caption{Lexical overlap --- finding the document sentence with the highest ROUGE against one reference sentence --- could be misleading. Examples are from the CNN/Daily Mail dataset~\cite{nallapati-etal-2016-abstractive}.}
    \label{table_bad}
    \vspace*{-.1cm}
    \end{table}

In this paper, we argue that the information coverage in summarization can be better evaluated by \textit{facet overlap}, \ie, whether the system summary covers the facets in the reference summary.
Specifically, we treat each \textit{reference sentence} as a facet, identify \textit{document sentences} that express the semantics of each facet as \textit{support sentences} of the facet, and measure information coverage by Facet-Aware Recall (\textbf{FAR}), \ie, how many facets are covered.
We focus on extractive summarization for the following two reasons.
Theoretically, since extractive methods cannot paraphrase or compress the document sentences as abstractive methods, it is somewhat unfair to penalize them for extracting long sentences that cover the facets.
Pragmatically, we can evaluate extractive methods automatically by comparing the indices of extracted sentences and support sentences.
We denote the mappings from each facet (sentence) in the reference summary to its support sentences in the document as Facet-Aware Mappings (\textbf{FAMs}).
FAMs can be used as labels indicating which sentences should be extracted but they are grouped with respect to each facet, while conventional extractive labels correspond to the entire reference summary rather than individual facets (detailed explanations in Sec.~\ref{sec_fam}). 
Compared to treating one summary as a sequence of n-grams, \textit{facet-aware evaluation} considers information coverage at a semantically richer granularity, and thus can contribute to a more accurate assessment on the summary quality.

To verify the effectiveness of facet-aware evaluation, we construct an \textit{extractive} version of the CNN/Daily Mail dataset~\cite{nallapati-etal-2016-abstractive} by annotating its FAMs (Sec.~\ref{sec_dataset}).
We revisit state-of-the-art extractive methods using this new extractive dataset (Sec.~\ref{sec_revisit}), the results of which show that FAR correlates better with human evaluation than ROUGE.
We also demonstrate that FAMs are beneficial for fine-grained evaluation of both abstractive and extractive methods (Sec.~\ref{sec_finegrained}).
We then illustrate how facet-aware evaluation can be useful for comparing different extractive methods in terms of their capability of extracting salient and non-redundant sentences (Sec.~\ref{sec_compare}).
Finally, we explore the feasibility of automatic FAM creation by evaluating sentence regression approaches against the ground-truth annotations (\ie, FAMs), and generalize facet-aware evaluation to the entire CNN/Daily Mail dataset \textit{without any human annotation} (Sec.~\ref{sec_AutoFAR}).
We believe that the summarization community will benefit from the proposed setup for better assessment of information coverage and gain deeper understandings of the current benchmark dataset and state-of-the-art methods through our analysis.

\start{Contributions.}
(1) We propose a facet-aware evaluation setup that better assesses information coverage for extractive summarization.
(2) We build the first dataset designed specifically for extractive summarization by creating facet-aware mappings from reference summaries to documents.
(3) We revisit state-of-the-art summarization methods in the proposed setup and discover valuable insights.
(4) To our knowledge, our work is also the first thorough quantitative analysis regarding the characteristics of the CNN/Daily Mail dataset.

\begin{figure}[t]
    \centering
    \includegraphics[width=1\linewidth]{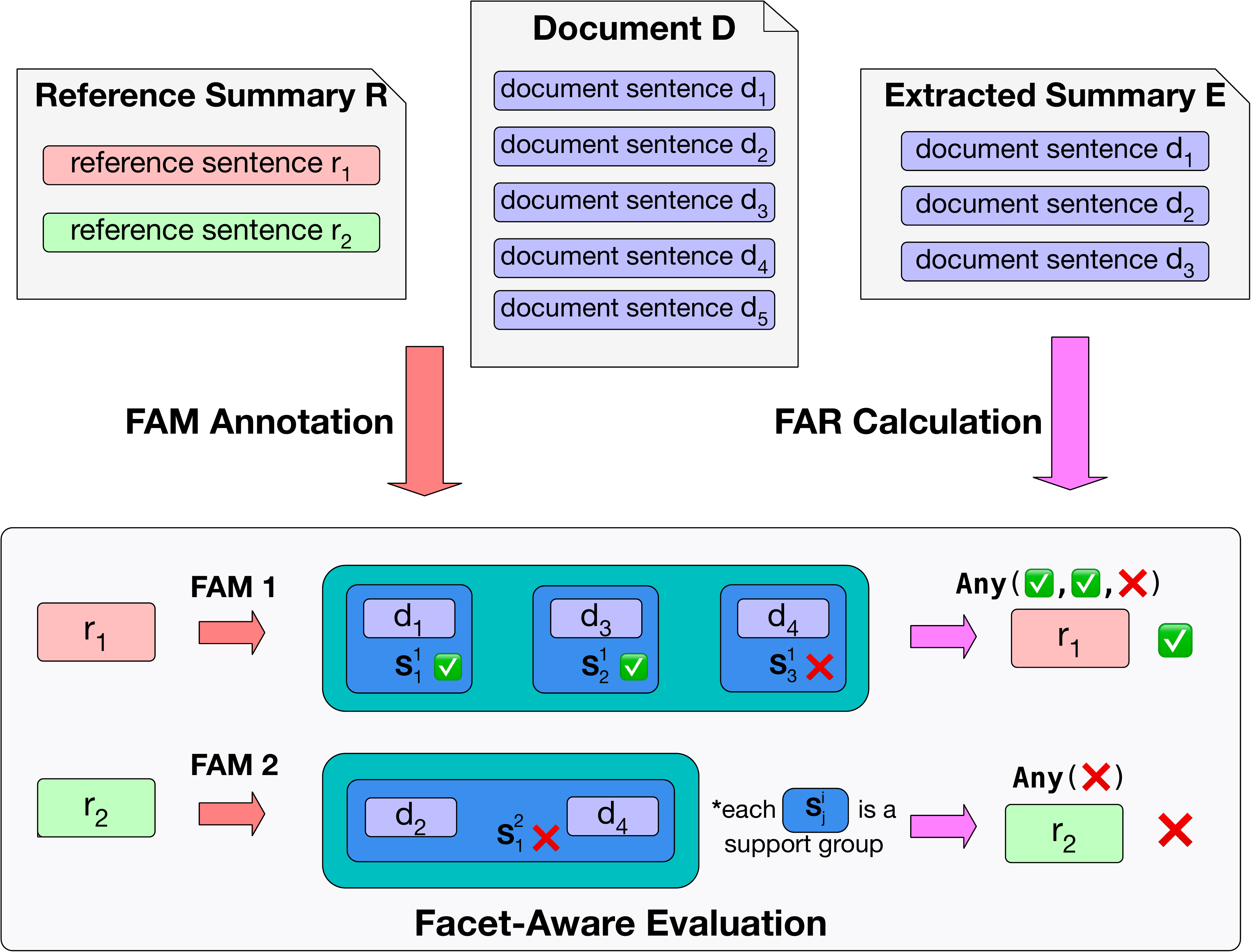}
    \vspace{-.1cm}
     \caption{\textbf{An illustration of facet-aware evaluation}. Two of three support groups of facet 1 ($r_1$) are covered. Facet 2 ($r_2$) cannot be covered as document sentence 4 ($d_4$) is missing in the extracted summary. The illustration corresponds to the example in Sec.~\ref{sec_far}.}
    \label{fig_far}
    \vspace{-.0cm}
\end{figure}

\section{Dataset Creation}
\label{sec_dataset}

\begin{table*}[t]
\small
    \centering
    \scalebox{.8}{
    \begin{tabular}{m{2.7cm} m{1.4cm}m{1.6cm}m{13.cm}}
        \toprule
         \textbf{Category} & \textbf{\#Samples} & \quad \textbf{\#Facets} & \quad \quad \quad   \textbf{Example} (full documents, reference summaries, and the FAMs can be found in Appendix~\ref{app:example})\\
        \midrule
         Noise (\textbf{N}) & 41 (27.3\%) & 137 (27.1\%) & 
        \vspace{-.3cm}
        \begin{itemize}
        \setlength\itemsep{-.5em}
            \item \textbf{Reference}: ``Furious 7'' opens Friday. (\textbf{unimportant detail})
             \item \textbf{Reference}: Click here for all the latest Floyd Mayweather vs Manny Pacquiao news. (\textbf{not found in the document}) 
            \item \textbf{Reference}: Vin Diesel: ``This movie is more than a movie''. (\textbf{random quotation})
            \item \textbf{Reference}: ``I had a small moment of awe,'' she said. (\textbf{random quotation})
            \vspace{-.3cm}
        \end{itemize}
\\
         \midrule
         Low Abstraction (\textbf{L}) & 89 (59.3\%) & 310 (61.2\%) $\overline{M}$=1: \quad \quad 275 (88.7\%) $\overline{M}$=2: \quad \quad 35 (11.3\%) & 
         
        \vspace{-.3cm}
        \begin{itemize}
        \setlength\itemsep{-.5em}
            \item \textbf{Reference}:  Willis never trademarked her most-famous work, calling it ``my gift to the city''. 
             \item \textbf{Support}: Willis never trademarked her most-famous work, calling it ``my gift to the city.'' (\textbf{identical})
            
            \vspace{.2cm}
            \item  \textbf{Reference}: Thomas K. Jenkins, 49, was arrested last month by deputies with the Prince George's County sheriff's office, authorities said.
            \item \textbf{Support}: Authorities said in a news release Thursday that 49-year-old Thomas K. Jenkins of capitol heights, Maryland, was arrested last month by deputies with the Prince George's County sheriff's office. (\textbf{compression})
            \vspace{-.3cm}
        \end{itemize}

\\
         \midrule
         High Abstraction (\textbf{H}) & 20 (13.3\%) & 59 (11.7\%) &
        
        \vspace{-.1cm}
        \begin{itemize}
        \setlength\itemsep{-.5em}
        \item \textbf{Reference}: College-bound basketball star asks girl with down syndrome to high school prom. Pictures of the two during the ``prom-posal'' have gone viral. (\textbf{highly abstractive})
         \item         
         \textbf{Reference}: While Republican Gov. Asa Hutchinson was weighing an Arkansas religious freedom bill, Walmart voiced its opposition. Walmart and other high-profile businesses are showing their support for gay and lesbian rights. (\textbf{unable to find support sentences})
         \vspace{-.1cm}
         \end{itemize}
         \\
        \bottomrule
        \end{tabular}
    }
    \vspace*{-.1cm}
    \caption{\textbf{Category breakdown of Facet-Aware Mappings (FAMs)}. Nearly 60\% samples are of low abstraction while more than a quarter of samples contain noisy facets. $\overline{M}$ denotes the average number of support sentences.}
    \label{table_data}
    \vspace*{-.0cm}
\end{table*}

In this section, we describe the process of creating an \textit{extractive} summarization dataset to facilitate facet-aware evaluation, which involves annotating FAMs between the documents and \textit{abstractive} reference summaries.
We first formalize the FAMs and then describe the FAM annotation on the CNN/Daily Mail dataset~\cite{nallapati-etal-2016-abstractive}.

\subsection{FAMs: Facet-Aware Mappings}
\label{sec_fam}
We denote one document-summary pair as $\{\mathbf{D}, \mathbf{R}\}$, where $\mathbf{D} = [d_1, d_2, ..., d_D]$, $\mathbf{R} = [r_1, r_2, ..., r_R]$, and $D$, $R$ denote the numbers of document sentences and reference sentences, respectively.
We conceptualize \textit{facet} as one unique semantic aspect presented in the summary.
In practice, we hypothesize that each reference sentence $r_i$ corresponds to one facet.\footnote{It is possible to define facet at sub-sentence or multi-sentence level as in Pyramid~\cite{nenkova-passonneau-2004-evaluating}. However, such definitions inevitably incur more annotation effort and lower inter-annotator agreement, while the current definition balances cost and effectiveness.}
We define \textit{support sentences} as the sentences in the document that express the semantics of one facet $r_i$.
We define \textit{support group} $\mathcal{S}$ of facet $r_i$ as a set of support sentences that can fully cover the information of $r_i$.
For each facet $r_i$ in the reference summary, we try to find all its support sentences in the document and put them into support groups.
Since we focus on single-document summarization in this work, most facets only have one support group. But some may contain multiple and extracting any of them would suffice (see example in Appendix~\ref{app:example} Table~\ref{table_multisup}). Allowing multiple support groups also makes FAMs easily extendable to multi-document summarization where redundant sentences prevail.

Formally, for each $r_i$, we annotate a Facet-Aware Mapping (FAM) $r_i \rightarrow \{ \mathcal{S}_1^i, \mathcal{S}_2^i, ..., \mathcal{S}_N^i \}$, where $N$ is the number of support groups.
Each $\mathcal{S}^i_j = \{d_{I_1}, d_{I_2}, ..., d_{I_{M_j}} \}$ is a support group, where  $I_1, I_2, ..., I_{M_j}$ are the indices of support sentences and $M_j$ is the number of support sentences in $\mathcal{S}^i_j$.
One illustrative example is presented in Fig.~\ref{fig_far}.
The support sentences are likely to be verbose, but we consider whether the support sentences express the semantics of the facet regardless of their length.\footnote{We ignore coreference (\eg, ``he" vs. ``the writer") and short fragments when considering the semantics of one facet, as we found that the wording of the reference summaries regarding such choices is also capricious.}
The reason is that we believe extractive summarization should focus on information coverage since it cannot alter the original sentences and once salient sentences are extracted, one can then compress them in an abstractive manner~\cite{chen-bansal-2018-fast,hsu-etal-2018-unified}.

 \start{Relation w. Extractive Labels.}
Extractive methods~\cite{nallapati2017summarunner,chen-bansal-2018-fast,narayan-etal-2018-ranking} typically require binary labels of every document sentence indicating whether it should be extracted during model training.
Such labels are called \textit{extractive labels} and usually created heuristically based on reference summaries since existing datasets do not provide extractive labels but only \textit{abstractive} references.
Our assumption that each reference sentence corresponds to one facet is similar to that during the creation of extractive labels.
The major differences are that 
(1) We allow an arbitrary number of support sentences while extractive labels usually limit to one support sentence for each reference sentence, \ie, we do not specify $M_j$. 
For example, we would put \textit{two} support sentences to \textit{one} support group if they are complementary and only combining them can cover the facet.
(2) We try to find multiple support groups ($N > 1$), as there could be more than one set of support sentences that cover the same facet.
In contrast, there is no notion of support group in extractive labels as they inherently form one such group ($N=1$).
Also, we allow $N=0$ if such a mapping cannot be found even by humans.
(3) The FAMs are more accurate as they are created by human annotators while extractive methods use sentence regression approaches (which we evaluate in Sec.~\ref{sec_sentLabel}) to obtain extractive labels approximately.

\start{Comparison w. SCUs.}
Some may mistake FAMs for Summarization Content Units (SCUs) in Pyramid~\cite{nenkova-passonneau-2004-evaluating}, but they are different in that (1) FAMs utilize both the documents and reference summaries while SCUs ignore the documents;
(2) FAMs are at the sentence level and can thus be used to \textit{automatically} evaluate extractive methods once created --- simply by matching sentence indices we can know how many facets are covered, while SCUs have to be \textit{manually} annotated for each system (refer to Appendix~\ref{sec_fig} Fig.~\ref{overview}).

\subsection{Creation of Extractive CNN/Daily Mail}
To verify the effectiveness of facet-aware evaluation, we annotate the FAMs of 150 document-summary pairs from the test set of CNN/Daily Mail.
Specifically, we take the first 50 samples in the test set, the 20 samples used in the human evaluation of~\citet{narayan-etal-2018-ranking}, and randomly draw another 80 samples.
The annotators are graduate students who are required to read through the document and mark support groups for each facet.
The most similar document sentences to each facet found by ROUGE and cosine similarity of average word embeddings are provided as the baselines for annotation.
310 non-empty FAMs are created by three annotators with high agreement (pairwise Jaccard index 0.714) and further verified to reach consensus.\footnote{One alternative way is to store multiple FAMs for each sample (like multiple reference summaries) and average their results as in ROUGE.}
On average, 5.44 (6.04 non-unique) document sentences are included as the support sentences in each document-summary pair.

To summarize, we found that the facets can be divided into three categories based on their quality and degree of abstraction as follows.

  \start{Noise}: The facet is noisy and irrelevant to the main content, either because the document itself is too hard to summarize (\eg, a report full of quotations) or the human editor was too subjective when writing the summary~\cite{see-etal-2017-get}. 
  Another possible reason is that the so-called ``summaries'' in CNN/Daily Mail are in fact ``story highlights'', which seems reasonable to include certain details. We found that 41/150 (27.3\%) samples have noisy facet(s), indicating that the reference summaries of CNN/Daily Mail are rather noisy.
  We show in Sec.~\ref{sec_revisit} that existing summarization methods perform poorly on this category, which justifies our judgment of ``noisy facets'' from another aspect.
  Also note that there would not be a ``noise'' category in a ``clean'' dataset. However, given the creation process of popular summarization datasets \cite{nallapati-etal-2016-abstractive,narayan-etal-2018-dont}, it is unlikely that all of their samples are of high quality.

  \start{Low Abstraction}: The facet can be mapped to its support sentences. 
  We denote the (rounded) average number of support sentences for each facet as $\overline{M}$ ($=\frac{1}{N}\sum_{j=1}^N M_j$, $N$ represents the number of support groups).
  As shown in Table~\ref{table_data}, all the facets with non-empty FAMs in CNN/Daily Mail are paraphrases or compression of one to two sentences in the document without much abstraction.

  \start{High Abstraction}: The facet \textit{cannot} be mapped to its support sentences ($N = 0$) by humans, which indicates that the writing of the facet requires deep understanding of the document rather than simply reorganizing several sentences.
  The proportion of this category (13.3\%) also indicates how often extractive methods would not work (well) on CNN/Daily Mail.

We found it easier than previously believed to create the FAMs on CNN/Daily Mail, as it is uncommon (average number of support groups $\overline{N} = 1.6$) to detect multiple sentences with similar semantics.
In addition, most support groups only have one or two support sentences with large lexical overlap, which coincides with the fact that extractive methods work quite well on CNN/Daily Mail and abstractive methods are often hybrid and learn to copy words directly from the documents.
That said, we try to automate the FAM creation and scale facet-aware evaluation to the whole test set of CNN/Daily Mail using machine-created FAMs (Sec.~\ref{sec_AutoFAR}).

\section{Facet-Aware Evaluation}

In this section, we introduce the facet-aware evaluation setup (Sec.~\ref{sec_far}) and demonstrate its effectiveness by revisiting state-of-the-art summarization methods under this new setup (Sec.~\ref{sec_revisit}). 
We then illustrate the additional benefits of facet-aware evaluation, including fine-grained evaluation (Sec.~\ref{sec_finegrained}) and comparative analysis (Sec.~\ref{sec_compare}).

\subsection{Proposed Metrics}
\label{sec_far}
As current extractive methods are facet-agnostic, \ie, their output is not nested (organized by facets) but a flat set of extracted sentences, we consider one facet as being ``covered" if any of its support groups can be found in the whole extracted summary.
Formally, we define the Facet-Aware Recall (FAR) as follows.
\begin{equation*}
    \text{FAR} = \frac{\sum_{i=1}^R \mathbf{Any}( \mathbf{I}(\mathcal{S}_1^i, \mathcal{E}), ..., \mathbf{I}(\mathcal{S}_N^i, \mathcal{E}) )}{R},
\end{equation*}
where $\mathbf{Any}(\mathcal{X})$ returns 1 if any $x \in \mathcal{X}$ is 1 and 0 otherwise, $\mathbf{I}(\mathcal{X}, \mathcal{Y})$ returns 1 if set $\mathcal{X} \subset \mathcal{Y}$ and 0 otherwise, $\mathcal{E}$ denotes the set of extracted sentences, and $R$ is the number of facets.
Intuitively, FAR does not over-penalize extractive methods for extracting long sentences as long as the extracted sentences cover the semantics of the facets.
FAR also treats each facet equally, whereas ROUGE weighs higher the facets with more tokens since they are more likely to incur lexical overlap.

To further measure model capability of retrieving salient (support) sentences without considering redundancy as FAR does, we merge all the support sentences of one document-summary pair to one single support set and define the Support-Aware Recall (SAR) as follows. 
SAR is used in Sec.~\ref{sec_compare} for the comparative analysis of extractive methods.
\begin{equation*}
    \text{SAR} = \frac{| \cup_{i=1}^R \cup_{j=1}^N  \mathcal{S}^i_j \cap \mathcal{E} |}{| \cup_{i=1}^R \cup_{j=1}^N  \mathcal{S}^i_j  |}.
\end{equation*}

\start{Example (Fig.~\ref{fig_far}).}
Assume that $R=2$, $r_1 \rightarrow \{ \{d_1\}, \{d_3 \}, \{d_4 \} \}, r_2 \rightarrow \{ \{d_2, d_4\} \}$, and $\mathcal{E} = \{d_1, d_2, d_3 \}$.
Then $\text{FAR} = \frac{1}{2}$ as $\mathcal{E}$ covers $\{d_1\}$ (or $\{d_3\}$) for $r_1$ but cannot cover $\{ d_2, d_4 \}$ for $r_2$.
$\text{SAR} = \frac{|\{d_1, d_2, d_3, d_4 \} \cap \{d_1, d_2, d_3 \}|}{|\{d_1, d_2, d_3, d_4 \}|} = \frac{3}{4}$.
Note that $d_1$ and $d_3$ are salient (support sentences) and both considered positive in SAR, while they only contribute to the coverage of one facet in FAR.

\subsection{Automatic Evaluation with FAR}
\label{sec_revisit}

By utilizing the low abstraction category on the extractive CNN/Daily Mail dataset, we revisit extractive methods to evaluate how they perform on information coverage.
Specifically, we compare Lead-3 (that extracts the first three document sentences), FastRL(E) (E for extractive only)~\cite{chen-bansal-2018-fast}, BanditSum~\cite{dong-etal-2018-banditsum}, NeuSum~\cite{zhou-etal-2018-neural}, Refresh~\cite{narayan-etal-2018-ranking}, and UnifiedSum(E)~\cite{hsu-etal-2018-unified} using both ROUGE and FAR. 
For a fair comparison, each method extracts three sentences ($|\mathcal{E}| = 3$).\footnote{Extracting all the sentences results in a perfect FAR, which is expected as FAR measures recall. One can also normalize FAR by the number of extracted sentences.}

\start{Results on Neural Extractive Methods.}
As shown in Table~\ref{table_sota}, there is almost no discrimination among the last four methods under ROUGE-1 F1, and the rankings under ROUGE-1/2/L often contradict with each other.
The observations on ROUGE Precision/Recall are similar. We provide them as well as more comparative analysis under facet-aware evaluation in Sec.~\ref{sec_compare}.
For facet coverage, the upper bound of FAR by extracting 3 sentences (Oracle, given the ground-truth FAMs) is 84.8, much higher than all the compared methods.
The best performing extractive method under FAR is UnifiedSum(E), which indicates that it covers the most facets semantically.

\begin{table}[ht]
    \centering
    \scalebox{.71}{
    \begin{tabular}{l ccc c}
        \toprule
         \textbf{Method} & \textbf{ROUGE-1} & \textbf{ROUGE-2} & \textbf{ROUGE-L} & \textbf{FAR}\\
        \midrule
        Lead-3 & 41.9 & 19.6 & 34.8 & 50.6 \\
        FastRL(E)& 41.6 & 20.3 & 35.5 & 50.8 \\
        BanditSum & 42.7 & 20.2 & 35.8 & 44.7 \\
        NeuSum & 42.7 & \textbf{22.1} & 36.4 & 51.2 \\
        Refresh & 42.8 & 20.3 & \textbf{39.3} & 51.3\\
        UnifiedSum(E) & 42.6 & 20.7 & 35.5 & \textbf{54.8} \\
        \midrule
        Oracle & 53.8 & 32.1 & 48.1 & 84.8 \\
        \bottomrule
    \end{tabular}
    }
    \vspace*{-.1cm}
    \caption{Performance comparison of extractive methods under ROUGE F1 and Facet-Aware Recall (FAR).} 
    \label{table_sota}
    \vspace*{-.1cm}
\end{table}

\start{FAR's Correlation w. Human Evaluation.}
Although FAR is supposed to be favored as the FAMs are manually labeled and indicate accurately whether one sentence should be extracted (assuming the annotations are in high quality), to further verify that FAR correlates with human preference, we ask the annotators to rank the outputs of UnifiedSum(E), NeuSum, and Lead-3 and measure ranking correlation. As listed in Table~\ref{table_rank}, we observe that the method with the most 1st ranks in the human evaluation coincides with FAR. 
We also find that FAR has higher Spearman's coefficient $\rho$ than ROUGE (0.457 vs. 0.44).\footnote{We expect that one can observe larger gains on datasets with less lexical overlap than CNN/Daily Mail.}

\begin{table}[ht]
    \centering
    \scalebox{.9}{
    \begin{tabular}{lccc}
        \toprule
         \textbf{Method} & \textbf{1st} & \textbf{2nd} & \textbf{3rd}\\
        \midrule
        Lead-3 & 26.8\% & 46.3\% & 26.8\% \\
        NeuSum & 29.3\% & 39.0\% & 31.7\% \\
        UnifiedSum(E) & \textbf{37.8\%} & \textbf{52.4\%} & 9.8\% \\
        \bottomrule
    \end{tabular}
    }
    \vspace*{-.1cm}
    \caption{\textbf{Proportions of system ranking in human evaluation}. FAR shows better human correlation than ROUGE and prefers UnifiedSum(E).} 
    \label{table_rank}
    \vspace*{-.1cm}
\end{table}

\subsection{Fine-grained Evaluation}
\label{sec_finegrained}

One benefit of facet-aware evaluation is that we can employ the \textit{category breakdown} of FAMs for fine-grained evaluation, namely, how one method performs on noisy / low abstraction / high abstraction samples, respectively.
Any metric of interest can be used for this fine-grained analysis. 
Here we consider ROUGE and additionally evaluate several abstractive methods: PG (Pointer-Generator)~\cite{see-etal-2017-get}, FastRL(E+A) (extractive+abstractive)~\cite{chen-bansal-2018-fast}, and UnifiedSum(E+A)~\cite{hsu-etal-2018-unified}.

As shown in Table~\ref{table_breakdown}, extractive methods perform poorly on high abstraction samples, which is somewhat expected since they cannot perform abstraction. Abstractive methods, however, also exhibit a huge performance gap between low and high abstraction samples, which suggests that existing abstractive methods achieve decent overall performance mainly by extraction rather than abstraction, \ie, performing well on low abstraction samples of CNN/Daily Mail.
We also found that all the compared methods perform much worse on the documents with ``noisy'' reference summaries, implying that the randomness in the reference summaries might introduce noise to both model training and evaluation. 
Note that although the sample size is relatively small, we observe consistent results when analyzing different subsets of the data.

\begin{table}[ht]
    \centering
    \scalebox{.82}{
    \begin{tabular}{c l cccc}
        \cmidrule[0.08em]{2-6}
         &\textbf{Method} & \textbf{N} & \textbf{L} & \textbf{H} & \textbf{L + H}\\
        \cmidrule{2-6}
        { \multirow{6}{*}{\rotatebox[origin=c]{90}{\small Extractive}}}
        & Lead-3 & 34.1 & 41.9 & 24.9 & 38.9\\
        &FastRL(E)& 33.5 & 41.6 & 31.2 & 39.8\\
        &BanditSum & 35.3 & 42.7 & \textbf{34.1} & \textbf{41.2} \\
        &NeuSum & 34.9 & 42.7 & 30.7 & 40.6\\
        &Refresh & \textbf{35.7} & 42.8 & 32.2 & 40.9\\
        &UnifiedSum(E) & 34.2 & 42.6 & 31.3 & 40.6\\

        \cmidrule{2-6}
        {\multirow{3}{*}{\rotatebox[origin=c]{90}{\small Abstractive}}}
        &PG & 32.6 & 40.6 & 27.5 & 38.2\\
        &FastRL(E+A)& 35.1 & 40.8  & 29.9& 38.8\\
        &UnifiedSum(E+A) & 34.2 & 42.4 & 29.2 & 40.1\\
        \cmidrule[0.08em]{2-6}
    \end{tabular}
    }
        \vspace*{-.1cm}
    \caption{ROUGE-1 F1 of extractive and abstractive methods on noisy (N), low abstraction (L), high abstraction (H), and high quality (L + H) samples.} 
    \label{table_breakdown}
    \vspace*{-.1cm}
\end{table}

\subsection{Comparative Analysis}
\label{sec_compare}

\begin{figure*}[th]
    \centering
    \includegraphics[width=.83\linewidth, height=10cm]{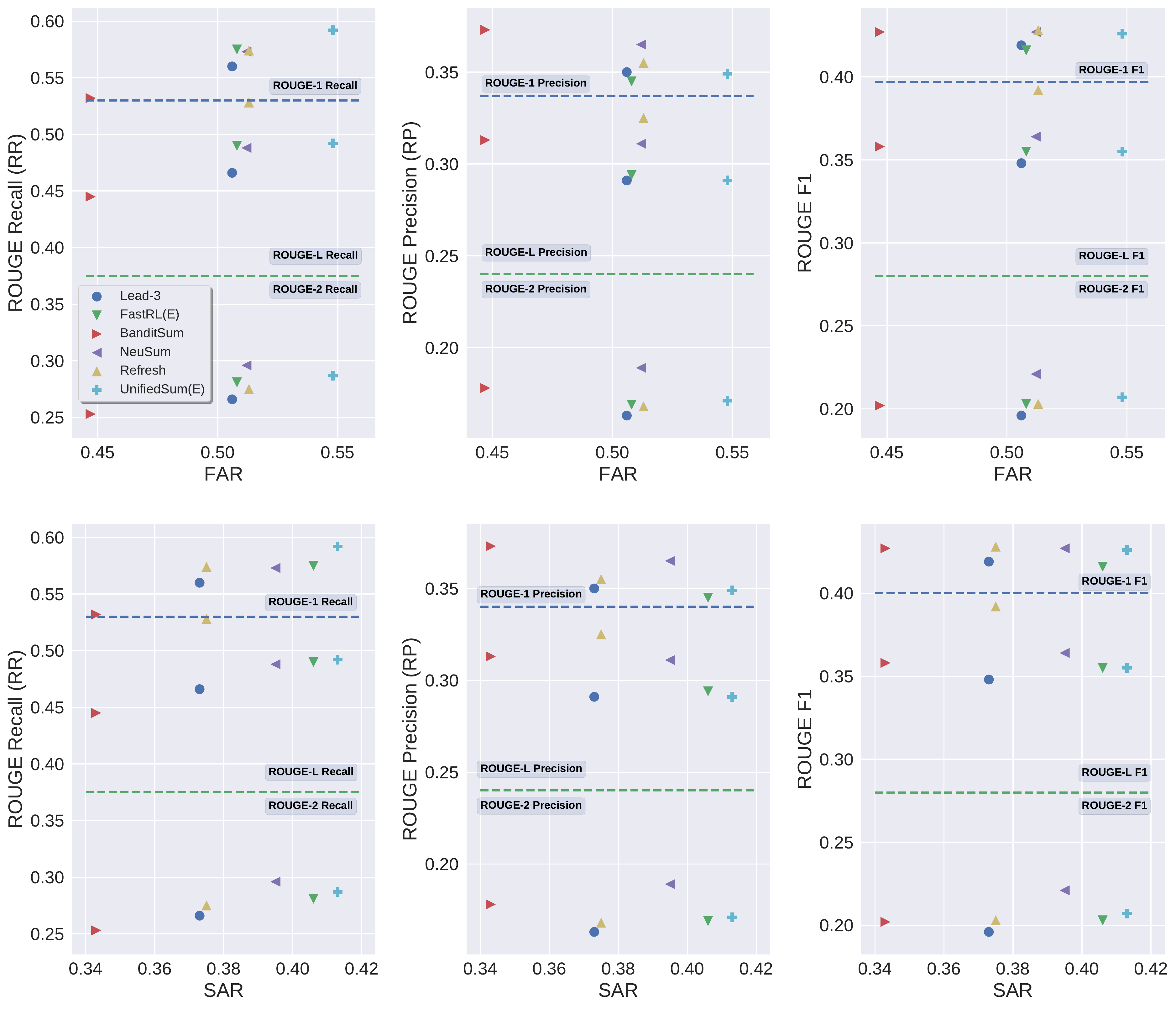}
    \vspace{-.1cm}
     \caption{\textbf{Performance of extractive methods under ROUGE, FAR, and SAR}. The results under ROUGE-1/2/L often disagree with each other. UnifiedSum(E) generally performs the best in the facet-aware evaluation.}
    \label{sar-far-all}
    \vspace{-.1cm}
\end{figure*}

Facet-aware evaluation is also beneficial for comparing extractive methods regarding their capability of extracting salient and non-redundant sentences.
We show the FAR, SAR, and ROUGE scores of various extractive methods in Fig.~\ref{sar-far-all}.
We next illustrate how one can leverage these scores under different metrics for comparative analysis. 
For brevity, we denote ROUGE Precision and ROUGE Recall as \textbf{RP} and \textbf{RR}, respectively.

\begin{figure}[th]
    \centering
    \includegraphics[width=.8\linewidth]{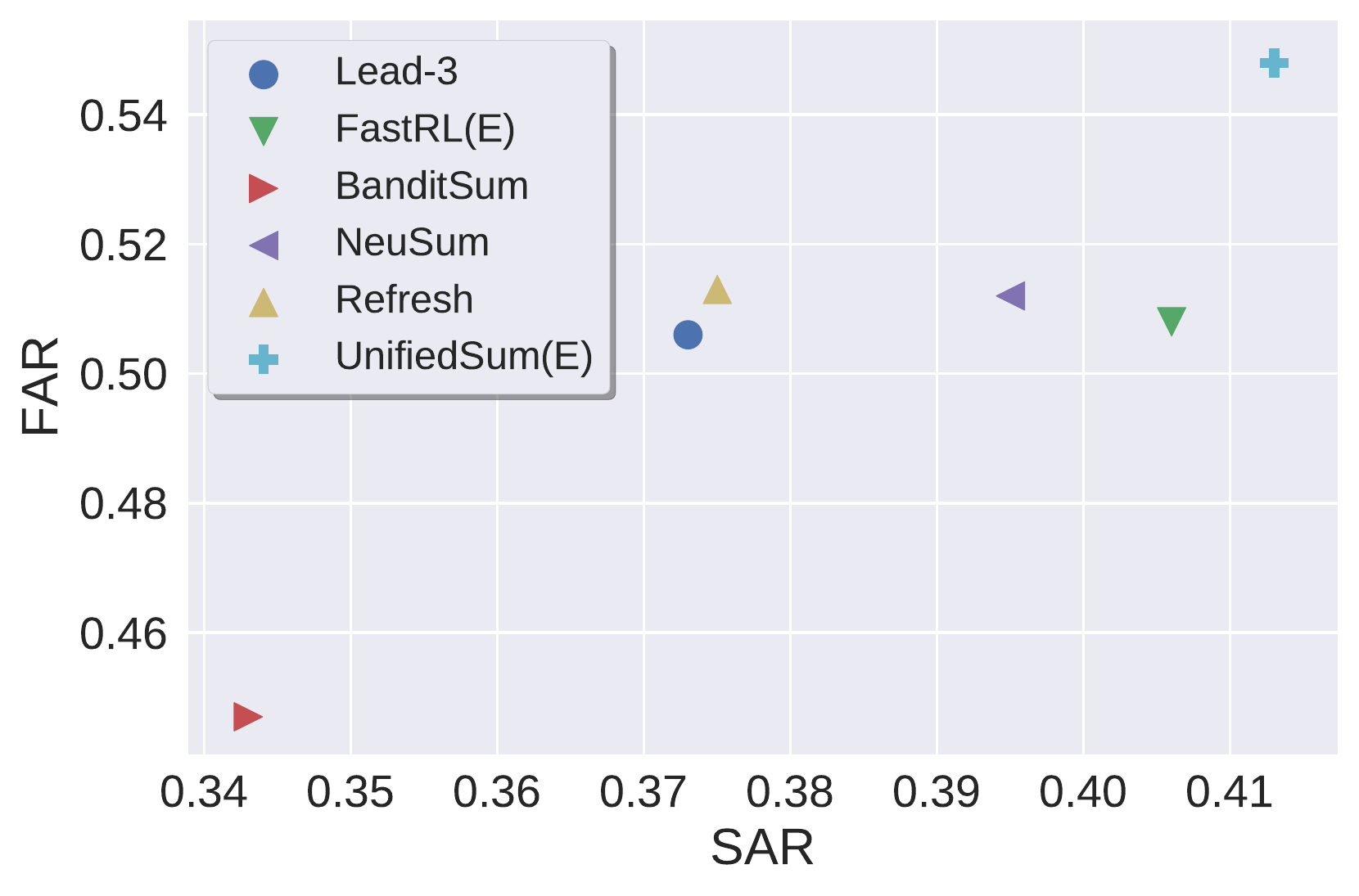}
    \vspace{-.1cm}
     \caption{Comparison of extractive methods under FAR and SAR reflects their capability of extracting salient and non-redundant sentences.}
    \label{sar-far}
    \vspace{-.1cm}
\end{figure}

\start{FAR vs. ROUGE.}
By comparing the scores of extractive methods under FAR and ROUGE, one can discover useful insights.
For example, we observe that the performance of Refresh, FastRL(E), NeuSum are quite close to Lead-3 under FAR, but they generally have higher RR.
Such results imply that these methods might have learned to extract sentences that are not the support sentences, \ie, sentences that do not directly contribute to the facet coverage, but still have lexical overlap with reference summaries.
It is also likely that they extract redundant support sentences that happen to have token matches with other facets.
Overall, UnifiedSum(E) covers the most facets (high FAR) and also has decent lexical matches (high RR).

\start{SAR vs. ROUGE.}
By comparing SAR with RP, one can find that UnifiedSum(E) extracts salient but possibly redundant support sentences, as it has higher SAR but similar RP to Lead-3.
On the contrary, Refresh has similar SAR with Lead-3 but higher RP, which again implies that it might extract non-support sentences that contain token matches but irrelevant semantics.
Similarly, BanditSum is capable of lexical overlap (high RP), but the matched tokens may not contribute much to the major semantics (low SAR).

\start{FAR vs. SAR.}
By comparing FAR with SAR (Fig.~\ref{sar-far}), we observe that FastRL(E) and NeuSum have FAR scores similar to Lead-3 and Refresh, but higher SAR scores.
One possible explanation is that FastRL(E) and NeuSum are better at extracting support sentences, but they do not handle redundancy very well, \ie, the extracted sentences might contain multiple support groups of the same facet (recall the example in Sec.~\ref{sec_far}).
For instance, there are 30.3\% extracted summaries of FastRL(E) that can cover more than one support group of the same facet while there are 19.1\% for Lead-3.

\section{Evaluation without Human Annotation}
\label{sec_AutoFAR}

In the previous sections, we have demonstrated the effectiveness and benefits of facet-aware evaluation.
One remaining issue that might prevent facet-aware evaluation from scaling is the need of human-annotated FAMs.
We thus study the feasibility of automatic FAM creation with sentence regression and present a pilot study of conducting facet-aware evaluation \textit{without any human annotation} in this section.

\subsection{Sentence Regression for FAM Creation}
\label{sec_sentLabel}

Similar to most benchmark constructions, facet-aware evaluation requires one-time annotation --- once the FAMs are annotated, we can reuse them for automatic evaluation.
That said, we explore various approaches to automate this one-time process. 
Specifically, we investigate whether facet-aware evaluation can be conducted without any human effort by utilizing \textit{sentence regression}~\cite{zopf-etal-2018-scores} to automatically create the FAMs.

Sentence regression is widely used to create extractive labels.
Sentence regression approaches typically transform abstractive reference summaries to extractive labels heuristically using ROUGE.
Previously, one could only estimate the quality of these labels by evaluating the extractive models trained using such labels, \ie, comparing their extracted summaries with the reference summaries (also approximately via ROUGE).
Now that the human-annotated FAMs serve as ground-truth extractive labels, we can evaluate how well each approach performs accurately.

\start{Sentence Regression Approaches.}
We briefly review recent sentence regression approaches as follows.
\citet{nallapati2017summarunner} greedily select sentences that maximize ROUGE-1 F1 until adding another sentence decreases it.
\citet{chen-bansal-2018-fast} find for each reference sentence the most similar sentence in the document by ROUGE-L recall.
\citet{zopf-etal-2018-scores} argue that precision is a better measure than recall because it aims not at covering as much information but at wasting as little space as possible.
\citet{narayan-etal-2018-ranking} measure sentence similarity by the average of ROUGE-1/2/L F1.
We also test other variants of ROUGE and TF-IDF, which represents sentences by TF-IDF features and measures their cosine similarity.

\subsection{Evaluation with Machine-Created FAMs}
\start{Results on Support Sentence Discovery.}
We first evaluate sentence regression with its original function, \ie, creating extractive labels (finding support sentences).
We merge the support groups of each sample and calculate precision and recall (\ie, SAR).
The performance of sentence regression approaches is shown in Table~\ref{table_approximate}.
The relatively low recall suggests that simply finding one support sentence for each facet as most existing approaches do would miss plenty of salient sentences, which could possibly worsen the models trained on such labels since the models would treat missed support sentences as unimportant ones.
On the bright side, many sentence regression approaches achieve high precision. For instance, 90.0\% document sentences labeled positive by \citet{narayan-etal-2018-ranking} indeed contain salient information.
This is to some extent explainable as ROUGE captures lexical overlap and as we have shown, there are many copy-and-paste reference summaries in CNN/Daily Mail.

\begin{table}[ht]
    \centering
    \scalebox{.8}{
    \begin{tabular}{lccc}
        \toprule
         \textbf{Method} & \textbf{Precision} & \textbf{Recall} & \textbf{F1}\\
        \midrule
        Lead-3 & 61.0 & 33.7 & 43.4 \\
        Greedy ROUGE-1 F1 & 58.2 & 30.8 & 40.3 \\
        TF-IDF & 83.7 & 51.9 & 64.0 \\
        ROUGE-1 F1 & 88.9 & 53.1 & 66.5 \\
        ROUGE-2 F1 & 86.6 & 52.3 & 65.2 \\
        ROUGE-L Recall & 89.3 & 53.7& 67.1 \\
        ROUGE-L Precision & 77.2 & 45.5 & 57.2 \\
        ROUGE-L F1 & 87.8 & 53.5 & 66.5 \\
        ROUGE-AVG F1 & \textbf{90.0} & \textbf{53.9} & \textbf{67.4} \\
        \bottomrule
    \end{tabular}
    }
        \vspace*{-.1cm}
    \caption{\textbf{Performance of sentence regression approaches regarding support sentence discovery}. High precision and low recall are often observed.} 
    \label{table_approximate}
    \vspace*{-.1cm}
\end{table}

\start{Correlation w. Human-Annotated FAMs.}
We then explore the correlation between human-annotated and machine-created FAMs by evaluating extractive methods against both of them.
This time we extend to find for each facet multiple support sentences and put each support sentence into a separate support group.
We measure the correlation between estimated and ground-truth FAR by Pearson's $r$.
We measure the correlation between system rankings induced from estimated and ground-truth FAR by Spearman's $\rho$ and Kendall's $\tau$.
The detailed correlation results of representative approaches are listed in Table~\ref{table_corr}.
We observe that creating three support groups consistently shows the highest correlation for the same sentence regression approach.
Also, the FAMs created by ROUGE-1 F1 and ROUGE-AVG F1 have very high correlation with human annotation, indicating the usability and reliability of machine-created FAMs for system ranking.

\begin{table}[ht]
    \centering
    \scalebox{.54}{
    \begin{tabular}{l  ccc  ccc  ccc}
        \cmidrule[0.08em]{1-10}
       \multirow{2}{*}{\textbf{Method}}   &  & \textbf{$N=1$} &  &  & \textbf{$N=2$} & & & \textbf{$N=3$}&\\
      \cmidrule(r){2-4} \cmidrule(r){5-7} \cmidrule(r){8-10}
       & $r$ & $\rho$ & $\tau$ & $r$ & $\rho$ & $\tau$ & $r$ & $\rho$ & $\tau$\\
        \cmidrule{1-10}
        ROUGE-1 F1 & 70.5 & 37.1 & 33.3 & 72.0 & 71.4 & 60.0 & \textbf{88.4} & \textbf{94.3} & \textbf{86.7}\\
        ROUGE-2 F1 & 11.0 & 25.7 & 20.0 & 43.4 & 65.7 & 46.7 & 88.4 & 65.7 & 60.0\\
        ROUGE-L F1 & 34.0 & 54.3 & 46.7 & 37.5 & 42.9 & 20.0 & 62.3 & 42.9 & 46.7\\
        ROUGE-AVG F1 & 49.6 & 54.3 & 46.7 & 46.1 & 65.7 & 46.7 & 83.2 & 82.9 & 73.3\\

        \cmidrule[0.08em]{1-10}
        
        \end{tabular}
    }
    \vspace{-.1cm}
    \caption{Correlation between ground-truth and estimated FAR scores by Pearson's $r$, Spearman's $\rho$, and Kendall's $\tau$. $N$ denotes the number of support groups.}
    \label{table_corr}
    \vspace*{-.1cm}
\end{table}

\start{FAR Prediction.}
Despite the high correlation, we also find that the estimated FAR scores may vary in range compared to the ground-truth FAR.\footnote{The raw estimated FAR scores are provided in Appendix~\ref{sec_fig} Fig.~\ref{fig-appro} in the interest of space.}
Therefore, we further use the estimations of different sentence regression approaches to train a linear regression model to fit the ground-truth FAR (denoted as AutoFAR).
We then calculate the estimated FAR scores on the whole test set of CNN/Daily Mail and use the trained linear regressor to predict a (supposedly) more accurate FAR score (denoted as AutoFAR-L).
As shown in Table~\ref{table_autofar}, the fitting of AutoFAR is very close to the ground-truth FAR, and the system ranking on the large-scale evaluation under AutoFAR-L follows a similar trend to that under FAR with Spearman's $\rho = 54.3$.
On the other hand, although our preliminary analysis on AutoFAR-L shows promising results, we also note that since the human annotation on the whole test set is lacking, the reliability of such extrapolation is not guaranteed and we leave more rigorous study with a larger number of systems and samples as future work.  

\begin{table}[ht]
    \centering
    \scalebox{.625}{
    \begin{tabular}{l ccc | c}
        \toprule
         \textbf{Method} & \textbf{FAR} & \textbf{AutoFAR} & \textbf{AutoFAR-L} & \textbf{FAR vs. AutoFAR(-L)}\\
        \midrule
        BanditSum & 44.7 & 44.8 & 44.7 & Pearson's $r$ \\
        Lead-3 & 50.6 & 51.3 & 45.6 & \textbf{97.6} (42.9)\\
        FastRL(E)& 50.8 & 51.0 & 43.1 & Spearman's $\rho$\\
        NeuSum & 51.2 & 49.9 & 44.3 & 77.1 (54.3)\\
        Refresh & 51.3 & 51.7 & 46.2 & Kendall's $\tau$\\
        UnifiedSum(E) & \textbf{54.8} & \textbf{54.5} & \textbf{46.9} & 60.0 (46.7)\\
        \bottomrule
    \end{tabular}
    }
        \vspace*{-.1cm}
    \caption{\textbf{FAR prediction via linear regression}. AutoFAR(-L) denotes the results on the human-annotated subset (entire CNN/Daily Mail dataset).} 
    \label{table_autofar}
    \vspace*{-.1cm}
\end{table}

\section{Related Work}
\label{sec_related}

\start{Evaluation Metrics for Text Summarization.}
ROUGE~\cite{lin-2004-rouge} is the most widely used evaluation metric for text summarization.
Extensions of ROUGE include ROUGE-WE~\cite{ng-abrecht-2015-better} that incorporated word embedding into ROUGE, ROUGE 2.0~\cite{ganesan2018rouge} that considered synonyms, and ROUGE-G~\cite{shafieibavani-etal-2018-graph} that applied graph analysis to WordNet for lexical and semantic matching.
Nevertheless, these extensions did not draw enough attention as the original ROUGE and recent advances \cite{gu2020generating,zhang2019pegasus} are still primarily evaluated by the vanilla ROUGE.

Another popular branch is Pyramid-based metrics~\cite{nenkova-passonneau-2004-evaluating,yang2016peak}, which annotate and compare the Summarization Content Units (SCUs) in the summaries.
FAR is related to Pyramid and HighRES~\cite{hardy-etal-2019-highres} in that Pyramid employs the summaries to annotate SCUs and HighRES highlights salient text fragments in the documents, while FAR considers both the summaries and documents.

Beyond lexical overlap, embedding-based evaluation metrics~\cite{zhang2019bertscore,zhao2019moverscore,sun-nenkova-2019-feasibility,xenouleas-etal-2019-sum} are gaining more traction along with the dominance of pre-trained language models. One straightforward way to incorporate embedding-based metrics into FAR is to use them as similarity measures instead of the ROUGE-based approaches tested in Sec.~\ref{sec_sentLabel} for automatic FAM creation (\ie, finding support sentences for each facet by the scores of embedding-based metrics).
Such similarity measures are especially beneficial when the facet and its support sentences are not similar at the lexical level.

\start{Reflections on Text Summarization.}
There has been increasing attention and critique to the issues of existing summarization metrics~\cite{schluter-2017-limits}, methods~\cite{kedzie-etal-2018-content,shapira-etal-2018-evaluating}, and datasets~\cite{jung-etal-2019-earlier}.
Notably, \citet{kryscinski2019neural} conducted a comprehensive critical evaluation for summarization from various aspects.
\citet{zopf-etal-2018-scores} investigated sentence regression approaches in a manner similar to ours but they could only evaluate them approximately against ROUGE as no ground-truth labels (FAMs) existed.

\start{Annotation and Analysis.}
Many recent studies conduct human annotation or evaluation on text summarization and other NLP tasks to gain useful insights.
\citet{hardy-etal-2019-highres} annotated 50 documents to demonstrate the benefits of highlight-based summarization evaluation. 
Recent summarization methods~\cite{paulus2017deep,narayan-etal-2018-ranking,chen-bansal-2018-fast} generally sampled 50 to 100 documents for human evaluation in addition to ROUGE in light of its limitations.
\citet{chen-etal-2016-thorough,yavuz-etal-2018-takes} inspected 100 samples and analyzed their category breakdown for reading comprehension and semantic parsing, respectively.
We observed similar trends when analyzing different subsets of the FAMs, indicating that our findings are relatively stable. 
We thus conjecture that our sample size is sufficient to verify our hypotheses and benefit future research.

\section{Conclusion and Future Work}
We propose a facet-aware evaluation setup for better assessment of information coverage in extractive summarization.
We construct an extractive summarization dataset and demonstrate the effectiveness of facet-aware evaluation on this newly constructed dataset, including better human correlation on the assessment of information coverage, and the support for fine-grained evaluation as well as comparative analysis. We also evaluate sentence regression approaches and explore the feasibility of fully-automatic evaluation without any human annotation.
In the future, we will investigate multi-document summarization datasets such as DUC~\cite{paul2004introduction} and TAC~\cite{dang2008overview} to see whether our findings coincide when multiple references are provided.
We will also explore better sentence regression approaches for the use of both extractive summarization methods and automatic FAM creation.

\section*{Acknowledgement}
We thank Woojeong Jin and Jiaming Shen for the valuable feedback on the paper draft.
We thank anonymous reviewers for the constructive comments.
Research was sponsored in part by US DARPA KAIROS Program No. FA8750-19-2-1004 and SocialSim Program No.  W911NF-17-C-0099, National Science Foundation IIS 16-18481, IIS 17-04532, and IIS-17-41317, and DTRA HDTRA11810026. Any opinions, findings, and conclusions or recommendations expressed herein are those of the authors and should not be interpreted as necessarily representing the views, either expressed or implied, of DARPA or the U.S. Government.

\bibliography{acl2020,anthology_small}
\bibliographystyle{acl_natbib}

\appendix
\clearpage

\begin{figure}[t]
    \centering
    \includegraphics[width=1\linewidth]{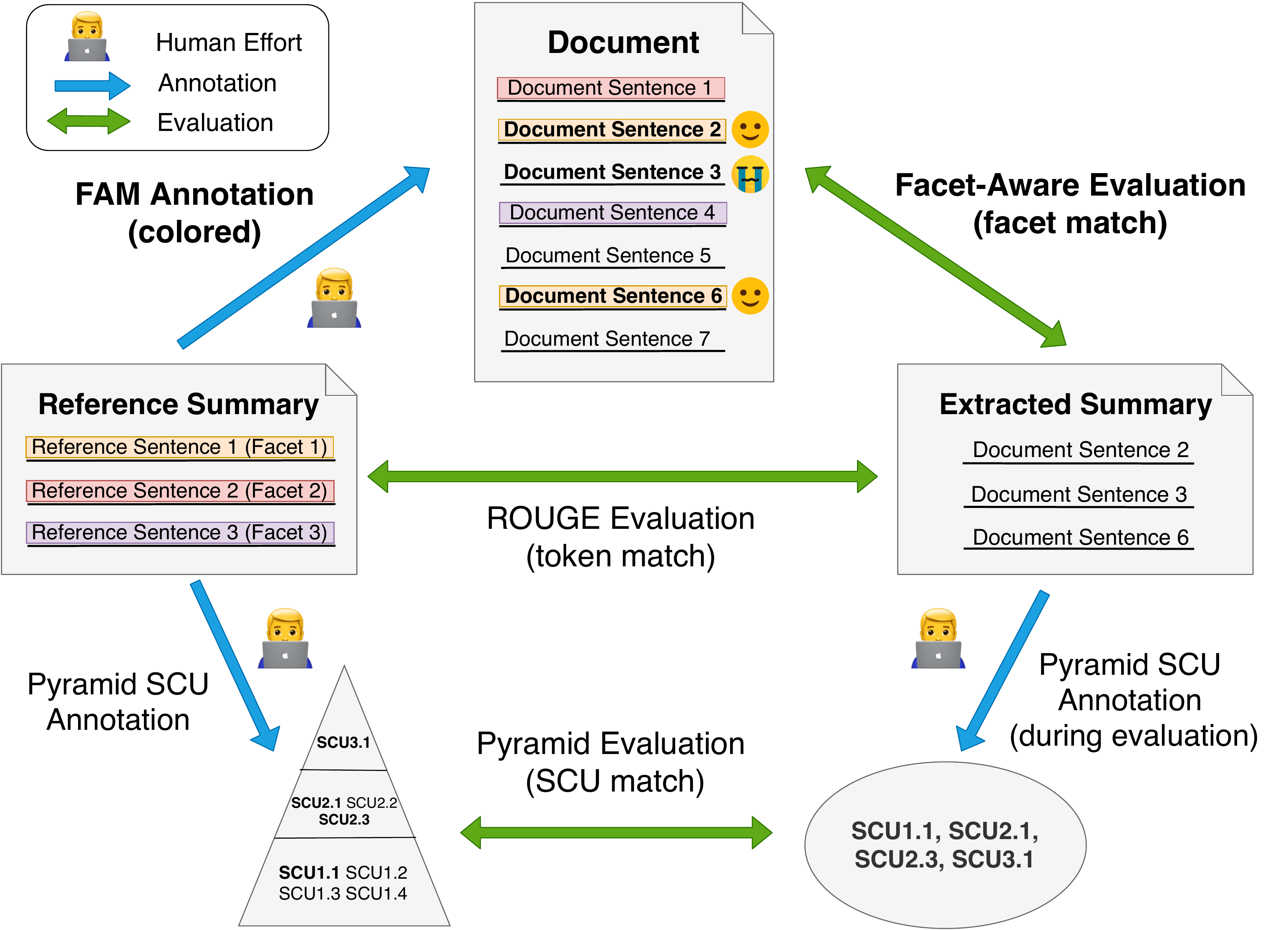}
    \vspace{-.25cm}
     \caption{\textbf{Comparison of summarization metrics}. Support sentences are marked in the same color as their corresponding facets. SCUs have to be annotated for each extracted summary during evaluation, while facet-aware evaluation can be conducted automatically by comparing sentence indices.}
    \label{overview}
    \vspace{-.35cm}
\end{figure}

\begin{figure*}[ht]
    \centering
    \includegraphics[width=1.0\linewidth]{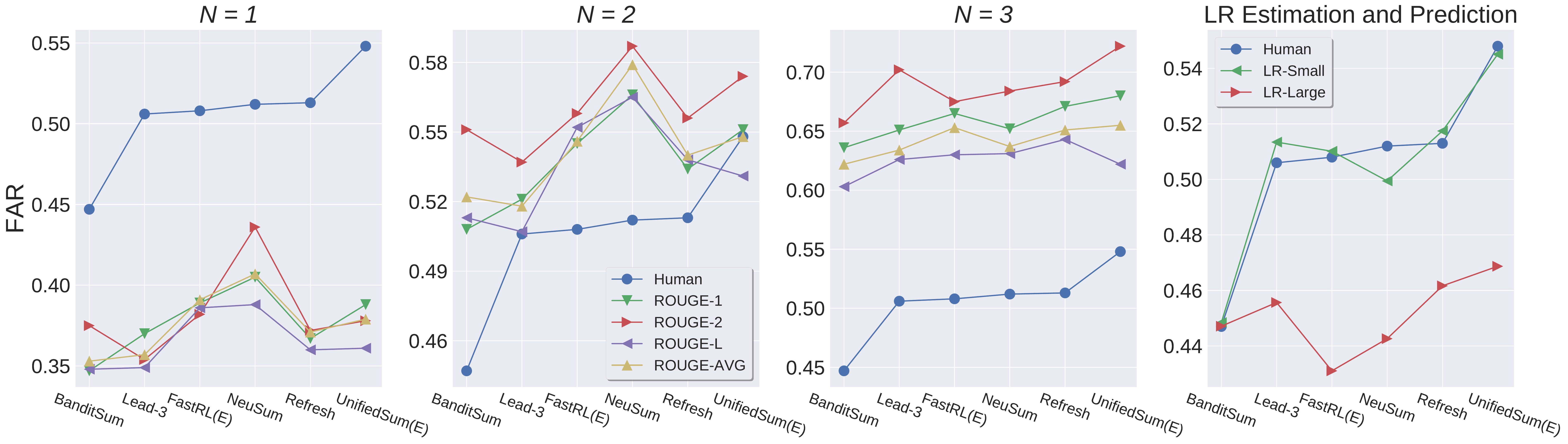}
    \vspace{-.4cm}
     \caption{The first three figures show the ground-truth and estimated FAR scores via human-annotated FAMs and machine-created FAMs. The fourth figure shows the fitting of linear regression on the human-annotated samples (LR-Small) and the prediction on the whole test set of CNN/Daily Mail (LR-Large). Systems are sorted in an ascending order by the ground-truth FAR on the human-annotated samples.}
    \label{fig-appro}
    \vspace{-.3cm}
\end{figure*}

\section{Practical Notes on CNN/Daily Mail}
We note several issues of the CNN/Daily Mail dataset in the hope that the researchers working on this dataset are better aware of these issues.

One issue is that sometimes the titles and image captions are introduced in the main body of the document by mistake (usually captured by ``-lrb- pictured -rrb-'' or colons), which may lead to bias or label leaking for model training since the reference summaries are observed to be similar to the titles and image captions~\cite{narayan-etal-2018-document}.
For example, we found that if there is a sentence in the main body that is almost the same as one of the captions, then that sentence is very likely to be used in the reference summary. Many such cases can be found in our annotated data.

We also found that in many documents, the 4-th sentence is ``\textit{scroll down for video}''. And if this sentence appears in one document, it is often the case that the first three sentences are good enough to summarize the whole document.
This finding provides yet another evidence why a simple Lead-3 baseline could be rather strong on CNN/Daily Mail.
In addition, sentences similar to the first three sentences can often be found afterward, which suggests that the first three sentences may not even belong to the main body of the document.

\section{Additional Illustration}
\label{sec_fig}
In Fig.~\ref{overview}, we show the comparison of ROUGE, FAR, and Pyramid.
In Fig.~\ref{fig-appro}, we show the the ground-truth FAR scores, the FAR scores estimated by various sentence regression approaches, and the prediction of FAR scores by linear regression.

\section{Detailed Examples}
\label{app:example}
We list below the full documents, reference summaries, and the corresponding FAMs of several examples shown in Table~\ref{table_data}.
In particular, Table \ref{table_multisup} shows an example of several support groups covering the same facet.
We release all of the annotated data to facilitate facet-aware evaluation and follow-up studies along this direction.

\begin{table*}[t]
\small
    \centering
    \scalebox{.8}{
    \begin{tabular}{p{20cm}}
        \toprule
         \textbf{ID}:
        1b2cc634e2bfc6f2595260e7ed9b42f77ecbb0ce \\
        \textbf{Category}:
        Noise \\
                 \midrule
        \textbf{Document}:\\
         -LRB- CNN -RRB- Paul Walker is hardly the first actor to die during a production .

But Walker 's death in November 2013 at the age of 40 after a car crash was especially eerie given his rise to fame in the `` Fast and Furious '' film franchise . The release of \textbf{`` Furious 7 '' on Friday} ({\color{red} \textbf{this is the only mention of ``Friday'' in the whole document}}) offers the opportunity for fans to remember -- and possibly grieve again -- the man that so many have praised as one of the nicest guys in Hollywood .

`` He was a person of humility , integrity , and compassion , '' military veteran Kyle Upham said in an email to CNN . Walker secretly paid for the engagement ring Upham shopped for with his bride .

`` We did n't know him personally but this was apparent in the short time we spent with him . I know that we will never forget him and he will always be someone very special to us , '' said Upham .

The actor was on break from filming `` Furious 7 '' at the time of the fiery accident , which also claimed the life of the car 's driver , Roger Rodas . Producers said early on that they would not kill off Walker 's character , Brian O'Connor , a former cop turned road racer .

Instead , the script was rewritten and special effects were used to finish scenes , with Walker 's brothers , Cody and Caleb , serving as body doubles .

There are scenes that will resonate with the audience -- including the ending , in which the filmmakers figured out a touching way to pay tribute to Walker while `` retiring '' his character .

At the premiere Wednesday night in Hollywood , Walker 's co-star and close friend \textbf{Vin Diesel gave a tearful speech before the screening , saying `` This movie is more than a movie . ''} ({\color{red} \textbf{random quotation, may use other quotes as well}})

`` You 'll feel it when you see it , '' Diesel said . `` There 's something emotional that happens to you , where you walk out of this movie and you appreciate everyone you love because you just never know when the last day is you 're gon na see them . ''

There have been multiple tributes to Walker leading up to the release . Diesel revealed in an interview with the `` Today '' show that he had named his newborn daughter after Walker . Social media has also been paying homage to the late actor .

A week after Walker 's death , about 5,000 people attended an outdoor memorial to him in Los Angeles . Most had never met him .

Marcus Coleman told CNN he spent almost \$ 1,000 to truck in a banner from Bakersfield for people to sign at the memorial .

`` It 's like losing a friend or a really close family member ... even though he is an actor and we never really met face to face , '' Coleman said . `` Sitting there , bringing his movies into your house or watching on TV , it 's like getting to know somebody . It really , really hurts . ''

Walker 's younger brother Cody told People magazine that he was initially nervous about how `` Furious 7 '' would turn out , but he is happy with the film .

`` It 's bittersweet , but I think Paul would be proud , '' he said .

CNN 's Paul Vercammen contributed to this report .
\\
\midrule
\textbf{Reference Summary}: \\
`` Furious 7 '' pays tribute to star Paul Walker , who died during filming

Vin Diesel : `` This movie is more than a movie '' ({\color{red}\textbf{random quotation}})

`` Furious 7 '' opens Friday ({\color{red}\textbf{unimportant detail}})
\\
\midrule
  \textbf{FAMs}: \\
  N/A\\
        \bottomrule
    \end{tabular}
    }
    \caption{Full document, reference summary, and the FAMs presented in Table~\ref{table_data}.} 
    
\end{table*}

\begin{table*}[t]
\small
     \centering
    \scalebox{.8}{
    \begin{tabular}{p{20cm}}
        \toprule
         \textbf{ID}:
        d58bf9387cd76f34bbb95fe25f8036015e5cc90a \\
        \textbf{Category}:
        Low Abstraction \\
                 \midrule
        \textbf{Document}:\\
         Dover police say a man they believe to be the so-called ` rat burglar ' who cut holes to tunnel into buildings has been arrested in Maryland .

{\color{purple}\textbf{Authorities said in a news release Thursday that 49-year-old Thomas K. Jenkins of Capitol Heights , Maryland , was arrested last month by deputies with the Prince George 's County Sheriff 's Office .}}

{\color{blue}\textbf{` Rat burglar ' : Thomas K. Jenkins , pictured is accused of robbing 18 Dover businesses}}

From September 2014 to February 2015 , Jenkins allegedly carried out 18 commercial robberies in Dover , Delaware , authorities there said .

` During the investigation it was learned that the Prince George 's County Sheriff 's Department had a series of burglaries that were similar in nature to the eighteen committed in Dover , ' the release said .

{\color{blue}\textbf{Thomas Jenkins has been accused by the Dover Police Department of robbing multiple businesses .}}

They are :

Maple Dale Country Club

Manlove Auto Parts

Sovereign Properties

Morgan Properties

U and I Builders

AMCO Check Cashing

Colonial Investment

1st Capital Mortgage

Advantage Travel

Ancient Way Massage

Tranquil Spirit Massage/Spa

Christopher Asay Massage

Morgan Communities

Vincenzo 's Restaurant

Happy Fortune Chinese Restaurant

Happy 13 Liquors

Del-One Credit Union

Pizza Time

Melvin 's Auto Service

Source : Dover Police Department/The News Journal

A car was found behind a building where a robbery took place and led deputies in Maryland to consider Jenkins as a suspect , authorities said .

Law enforcement later found Jenkin 's car and tracked where he went , Dover police said .

{\color{red} \textbf{Police say Jenkins had cut a hole in the roof of a commercial business in Maryland on March 9 and deputies arrested him as he fled .}}

According to Dover police , ` Jenkins was found in possession of .45 - caliber handgun that was stolen from a business in Delaware State Police Troop 9 jurisdiction . A search of Jenkins vehicle revealed an additional .45 - caliber handgun stolen from the same business . '

{\color{blue}\textbf{Jenkins is being held in Maryland and will face 72 charges involving the 18 burglaries in Dover when he is returned to Delaware .}}

The charges he is facing break down to : four counts of wearing a disguise during the commission of a felony , eighteen counts of third-degree burglary , eighteen counts of possession of burglary tools , fourteen counts of theft under \$ 1,500 , and eighteen counts of criminal mischief , two of which are felonies , authorities said .

{\color{orange}\textbf{Cpl. Mark Hoffman with the Dover Police Department told the News Journal that Delaware State Police are planning to file charges over a 19th robbery at Melvin 's Auto Service , which reportedly occurred in a part of Dover where jurisdiction is held by state police .}}

Sharon Hutchison , who works at one of the businesses Jenkins allegedly robbed , told the newspaper ` He cut through two layers of drywall , studs and insulation . '

The Prince George 's County Sheriff 's Department did not immediately return a request for information on what charges Jenkins is facing there .
\\
\midrule
\textbf{FAMs}: \\
\begin{itemize}

\item {\color{purple}\textbf{thomas k. jenkins , 49 , was arrested last month by deputies with the prince george 's county sheriff 's office , authorities said .}}

\textbf{[Support Group0][Sent0]}: authorities said in a news release thursday that 49-year-old thomas k. jenkins of capitol heights , maryland , was arrested last month by deputies with the prince george 's county sheriff 's office .

\item {\color{red}\textbf{police say jenkins had cut a hole in the roof of a commercial business in maryland on march 9 and deputies arrested him as he fled .}}

\textbf{[Support Group0][Sent0]}: police say jenkins had cut a hole in the roof of a commercial business in maryland on march 9 and deputies arrested him as he fled .

\item {\color{blue}\textbf{jenkins is accused of carrying out multiple robberies in dover , delaware .}}

\textbf{[Support Group0][Sent0]}: jenkins is being held in maryland and will face 72 charges involving the 18 burglaries in dover when he is returned to delaware .

\textbf{[Support Group1][Sent0]}: ` rat burglar ' : thomas k. jenkins , pictured is accused of robbing 18 dover businesses .

\textbf{[Support Group2][Sent0]}: thomas jenkins has been accused by the dover police department of robbing multiple businesses .

\item {\color{blue}\textbf{he is facing 72 charges from the dover police department for 18 robberies .}}

\textbf{[Support Group0][Sent0]}: jenkins is being held in maryland and will face 72 charges involving the 18 burglaries in dover when he is returned to delaware .

\item {\color{orange}\textbf{the delaware state police is planning to file charges over a 19th robbery , which occurred in a part of dover where jurisdiction is held by state police .}}

\textbf{[Support Group0][Sent0]}: mark hoffman with the dover police department told the news journal that delaware state police are planning to file charges over a 19th robbery at melvin 's auto service , which reportedly occurred in a part of dover where jurisdiction is held by state police .

\end{itemize}
\\
        \bottomrule
    \end{tabular}
    }
    \caption{Full document, reference summary, and the FAMs presented in Table~\ref{table_data}.} 
   \label{table_multisup}
\end{table*}

\begin{table*}[t]
\small
     \centering
    \scalebox{.8}{
    \begin{tabular}{p{20cm}}
        \toprule
         \textbf{ID}:
        d1fa0db909ce45fe1ee32d6cbb546e9d784bcf74 \\
        \textbf{Category}:
        Low Abstraction \\
                 \midrule
        \textbf{Document}:\\
         -LRB- CNN -RRB- You probably never knew her name , but you were familiar with her work .

Betty Whitehead Willis , the designer of the iconic `` Welcome to Fabulous Las Vegas '' sign , died over the weekend . She was 91 .

{\color{blue}\textbf{Willis played a major role in creating some of the most memorable neon work in the city .}}

The Neon Museum also credits her with designing the signs for Moulin Rouge Hotel and Blue Angel Motel

Willis visited the Neon Museum in 2013 to celebrate her 90th birthday .

Born about 50 miles outside of Las Vegas in Overton , she attended art school in Pasadena , California , before returning home .

She retired at age 77 .

{\color{red}\textbf{Willis never trademarked her most-famous work , calling it `` my gift to the city . ''}}

Today it can be found on everything from T-shirts to refrigerator magnets .

People we 've lost in 2015
\\
\midrule
\textbf{FAMs}: \\
\begin{itemize}

\item {\color{red}\textbf{willis never trademarked her most-famous work , calling it `` my gift to the city ''}}

\textbf{[Support Group0][Sent0]}: willis never trademarked her most-famous work , calling it `` my gift to the city . ''

\item {\color{blue}\textbf{she created some of the city 's most famous neon work .}}

\textbf{[Support Group0][Sent0]}: willis played a major role in creating some of the most memorable neon work in the city .

\end{itemize}
\\
        \bottomrule
    \end{tabular}
    }
    \caption{Full document, reference summary, and the FAMs presented in Table~\ref{table_data}.} 
   
\end{table*}

\begin{table*}[t]
\small
     \centering
    \scalebox{.8}{
    \begin{tabular}{p{20cm}}
        \toprule
         \textbf{ID}:
        dc833f8b55e381011ce23f89ea909b9a141b5a66 \\
        \textbf{Category}:
        High Abstraction \\
                 \midrule
        \textbf{Document}:\\
         -LRB- CNN -RRB- As goes Walmart , so goes the nation ?

Everyone from Apple CEO Tim Cook to the head of the NCAA slammed religious freedom laws being considered in several states this week , warning that they would open the door to discrimination against gay and lesbian customers .

But it was the opposition from Walmart , the ubiquitous retailer that dots the American landscape , that perhaps resonated most deeply , providing the latest evidence of growing support for gay rights in the heartland .

Walmart 's staunch criticism of a religious freedom law in its home state of Arkansas came after the company said in February it would boost pay for about 500,000 workers well above the federal minimum wage . Taken together , the company is emerging as a bellwether for shifting public opinion on hot-button political issues that divide conservatives and liberals .

And some prominent Republicans are urging the party to take notice .

Former Minnesota Gov. Tim Pawlenty , who famously called on the GOP to `` be the party of Sam 's Club , not just the country club , '' told CNN that Walmart 's actions `` foreshadow where the Republican Party will need to move . ''

`` The Republican Party will have to better stand for '' ideas on helping the middle class , said Pawlenty , the head of the Financial Services Roundtable , a Washington lobbying group for the finance industry . The party 's leaders must be `` willing to put forward ideas that will help modest income workers , such as a reasonable increase in the minimum wage , and prohibit discrimination in things such as jobs , housing , public accommodation against gays and lesbians . ''

Walmart , which employs more than 50,000 people in Arkansas , emerged victorious on Wednesday . Hours after the company 's CEO , Doug McMillon , called on Republican Gov. Asa Hutchinson to veto the bill , the governor held a news conference and announced he would not sign the legislation unless its language was fixed .

Walmart 's opposition to the religious freedom law once again puts the company at odds with many in the Republican Party , which the company 's political action committee has tended to support .

In 2004 , the Walmart PAC gave around \$ 2 million to Republicans versus less than \$ 500,000 to Democrats , according to data from the Center for Responsive Politics . That gap has grown less pronounced in recent years . In 2014 , the PAC spent about \$ 1.3 million to support Republicans and around \$ 970,000 for Democrats .

It has been a gradual transformation for Walmart .

In 2011 , the company bulked up its nondiscrimination policies by adding protections for gender identity . Two years later , the company announced that it would start offering health insurance benefits to same-sex partners of employees starting in 2014 .

Retail experts say Walmart 's evolution on these issues over the years is partly a reflection of its diverse consumer base , as well as a recognition of the country 's increasingly progressive views of gay equality -LRB- support for same-sex marriage is at a new high of 59 \% , according to a recent Wall Street Journal/NBC News poll -RRB- .

`` It 's easy for someone like a Chick-fil-A to take a really polarizing position , '' said Dwight Hill , a partner at the retail consulting firm McMillanDoolittle . `` But in the world of the largest retailer in the world , that 's very different . ''

Hill added : Same-sex marriage , `` while divisive , it 's becoming more common place here within the U.S. , and the businesses by definition have to follow the trend of their customer . ''

The backlash over the religious freedom measures in Indiana and Arkansas this week is shining a bright light on the broader business community 's overwhelming support for workplace policies that promote gay equality .

After Indiana Gov. Mike Pence , a Republican , signed his state 's religious freedom bill into law , CEOs of companies big and small across the country threatened to pull out of the Hoosier state .

The resistance came from business leaders of all political persuasions , including Bill Oesterle , CEO of the business-rating website Angie 's List and a one-time campaign manager for former Indiana Gov. Mitch Daniels . Oesterle announced that his company would put plans on hold to expand its footprint in Indianapolis in light of the state 's passage of the religious freedom act .

NASCAR , scheduled to hold a race in Indianapolis this summer , also spoke out against the Indiana law .

`` What we 're seeing over the past week is a tremendous amount of support from the business community who are standing up and are sending that equality is good for business and discrimination is bad for business , '' said Jason Rahlan , spokesman for the Human Rights Campaign .

The debate has reached presidential politics .

National Republicans are being forced to walk the fine line of protecting religious liberties and supporting nondiscrimination .

Likely GOP presidential candidate Jeb Bush initially backed Indiana 's religious freedom law and Pence , but moderated his tone a few days later . The former Florida governor said Wednesday that Indiana could have taken a `` better '' and `` more consensus-oriented approach . ''

`` By the end of the week , Indiana will be in the right place , '' Bush said , a reference to Pence 's promise this week to fix his state 's law in light of the widespread backlash .

Others in the GOP field are digging in . Sen. Ted Cruz of Texas , the only officially declared Republican presidential candidate , said Wednesday that he had no interest in second-guessing Pence and lashed out at the business community for opposing the law .

`` I think it is unfortunate that large companies today are listening to the extreme left wing agenda that is driven by an aggressive gay marriage agenda , '' Cruz said .

Meanwhile , former Secretary of State Hillary Clinton , who previously served on Walmart 's board of directors , called on Hutchinson to veto the Arkansas bill , saying it would `` permit unfair discrimination '' against the LGBT community .

Jay Chesshir , CEO of the Little Rock Regional Chamber of Commerce in Arkansas , welcomed Hutchinson 's pledge on Wednesday to seek changes to his state 's bill . He said businesses are not afraid to wade into a politically controversial debate to ensure inclusive workplace policies .

`` When it comes to culture and quality of life , businesses are extremely interested in engaging in debate simply because it impacts its more precious resource -- and that 's its people , '' Chesshir said . `` Therefore , when issues arise that have negative or positive impact on those things , then the business community will again speak and speak loudly . ''\\
\midrule
\textbf{Reference Summary}:\\
While Republican Gov. Asa Hutchinson was weighing an Arkansas religious freedom bill , Walmart voiced its opposition ({\color{red}\textbf{highly abstractive, hard to obtain by rephrasing original sentences}})

Walmart and other high-profile businesses are showing their support for gay and lesbian rights

Their stance puts them in conflict with socially conservative Republicans , traditionally seen as allies\\
\midrule
  \textbf{FAMs}: \\
  N/A\\
        \bottomrule
    \end{tabular}
    }
    \caption{Full document, reference summary, and the FAMs presented in Table~\ref{table_data}.} 
   
\end{table*}

\begin{table*}[t]
\small
     \centering
    \scalebox{.8}{
    \begin{tabular}{p{20cm}}
        \toprule
         \textbf{ID}:
        1b2cc634e2bfc6f2595260e7ed9b42f77ecbb0ce \\
        \textbf{Category}:
        High Abstraction \\
                 \midrule
        \textbf{Document}:\\
         -LRB- CNN -RRB- He 's a blue chip college basketball recruit . She 's a high school freshman with Down syndrome .

At first glance Trey Moses and Ellie Meredith could n't be more different . But all that changed Thursday when Trey asked Ellie to be his prom date .

Trey -- a star on Eastern High School 's basketball team in Louisville , Kentucky , who 's headed to play college ball next year at Ball State -- was originally going to take his girlfriend to Eastern 's prom .

So why is he taking Ellie instead ? `` She 's great ... she listens and she 's easy to talk to '' he said .

Trey made the prom-posal -LRB- yes , that 's what they are calling invites to prom these days -RRB- in the gym during Ellie 's P.E. class .

Trina Helson , a teacher at Eastern , alerted the school 's newspaper staff to the prom-posal and posted photos of Trey and Ellie on Twitter that have gone viral . She was n't surpristed by Trey 's actions .

`` That 's the kind of person Trey is , '' she said .

To help make sure she said yes , Trey entered the gym armed with flowers and a poster that read `` Let 's Party Like it 's 1989 , '' a reference to the latest album by Taylor Swift , Ellie 's favorite singer .

Trey also got the OK from Ellie 's parents the night before via text . They were thrilled .

`` You just feel numb to those moments raising a special needs child , '' said Darla Meredith , Ellie 's mom . `` You first feel the need to protect and then to overprotect . ''

Darla Meredith said Ellie has struggled with friendships since elementary school , but a special program at Eastern called Best Buddies had made things easier for her .

She said Best Buddies cultivates friendships between students with and without developmental disabilities and prevents students like Ellie from feeling isolated and left out of social functions .

`` I guess around middle school is when kids started to care about what others thought , '' she said , but `` this school , this year has been a relief . ''

Trey 's future coach at Ball State , James Whitford , said he felt great about the prom-posal , noting that Trey , whom he 's known for a long time , often works with other kids

Trey 's mother , Shelly Moses , was also proud of her son .

`` It 's exciting to bring awareness to a good cause , '' she said . `` Trey has worked pretty hard , and he 's a good son . ''

Both Trey and Ellie have a lot of planning to do . Trey is looking to take up special education as a college major , in addition to playing basketball in the fall .

As for Ellie , she ca n't stop thinking about prom .

`` Ellie ca n't wait to go dress shopping '' her mother said .

`` Because I 've only told about a million people ! '' Ellie interjected .
\\
\midrule
\textbf{Reference Summary}: \\
College-bound basketball star asks girl with down syndrome to high school prom. ({\color{red}\textbf{highly abstractive, hard to obtain by rephrasing original sentences}})

Pictures of the two during the ``prom-posal'' have gone viral.
\\
\midrule
  \textbf{FAMs}: \\
  N/A\\
        \bottomrule
    \end{tabular}
    }
    \caption{Full document, reference summary, and the FAMs presented in Table~\ref{table_data}.} 
   
\end{table*}

\end{document}